\documentclass[sigconf]{acmart}
\usepackage{fancyhdr}

\AtBeginDocument{%
  \providecommand\BibTeX{{%
    \normalfont B\kern-0.5em{\scshape i\kern-0.25em b}\kern-0.8em\TeX}}}

\copyrightyear{2021}
\acmYear{2021}
\setcopyright{acmcopyright}\acmConference[MMAsia '20]{ACM Multimedia
Asia}{March 7--9, 2021}{Virtual Event, Singapore}
\acmBooktitle{ACM Multimedia Asia (MMAsia '20), March 7--9, 2021, Virtual Event,
Singapore}
\acmPrice{15.00}
\acmDOI{10.1145/3444685.3446282}
\acmISBN{978-1-4503-8308-0/21/03}

\begin{document}
\fancyhead{}
\title{Full-Resolution Encoder–Decoder Networks with Multi-Scale Feature Fusion for Human Pose Estimation}



\author{Jie Ou}
\affiliation{%
  \institution{University of Electronic Science and Technology of China}
  \city{Chengdu}
  \country{China}}
\email{oujieww6@gmail.com}

\author{Mingjian Chen}
\affiliation{%
  \institution{University of Electronic Science and Technology of China}
  \city{Chengdu}
  \country{China}}
\email{starvingnow@163.com}

\author{Hong Wu}
\affiliation{%
  \institution{University of Electronic Science and Technology of China}
  \city{Chengdu}
  \country{China}}
\email{hwu@uestc.edu.cn}
\authornote{Corresponding authors}





\begin{abstract}
 
To achieve more accurate 2D human pose estimation, we extend the successful encoder-decoder network, simple baseline network (SBN), in three ways. To reduce the quantization errors caused by the large output stride size, two more decoder modules are appended to the end of the simple baseline network to get full output resolution. Then, the global context blocks (GCBs) are added to the encoder and decoder modules to enhance them with global context features. Furthermore, we propose a novel spatial-attention-based multi-scale feature collection and distribution module (SA-MFCD) to fuse and distribute multi-scale features to boost the pose estimation. Experimental results on the MS COCO dataset indicate that our network can remarkably improve the accuracy of human pose estimation over SBN, our network using ResNet34 as the backbone network can even achieve the same accuracy as SBN with ResNet152, and our networks can achieve superior results with big backbone networks.

\end{abstract}




\maketitle

\section{Introduction}
2D human pose estimation, which aims to predict locations of human body joints in an image, is a very important and challenging computer vision task. It has been studied for decades and the traditional methods mainly rely on hand-craft features~\cite{liu2015survey}. With the application of convolutional neural networks, the accuracy of human pose estimation has been greatly improved~\cite{wei2016convolutional,newell2016stacked,chen2018cascaded,moon2019posefix,xiao2018simple}. 

Most multi-person pose estimation methods can be divided into two categories: top-down and bottom-up methods. The top-down methods~\cite{papandreou2017towards,fang2017rmpe,xiao2018simple,sun2019deep} first use a human detector to detect all person instances in an image, then perform single-person pose estimation for each of them respectively. The top-down methods is inefficient in case of crowds of persons because the run-time of the second step is proportional to the number of persons. On the contrary, the bottom-up methods~\cite{papandreou2018personlab,pishchulin2016deepcut,insafutdinov2017arttrack,cao2017realtime} detect the joints of all persons at once and then group the joints into each individual person by a grouping algorithm. The bottom-up methods suffer from high complexity of joint grouping step and the multi-scale problem in joint detection. Recently, most state-of-the-art results have been achieved by the top-down methods. In this paper, we follow the top-down strategy and develop an effective human pose estimation method.  

Among the top-down methods, the simple baseline network~\cite{xiao2018simple} has achieved state-of-the-art results with a simple encoder-decoder architecture. Due to its simplicity and effectiveness, it can be used as a basic network to develop more advanced pose estimation methods. To achieve more accurate human pose estimation, we extend the simple baseline network in three ways. To reduce the quantization error caused by large output stride size, two more decoder modules are appended to the end of SBN to increase the output resolution to the same as input resolution. Then, the global context blocks (GCBs) are added to the encoder and decoder modules to enhance them with global context features. Furthermore, we develop a novel spatial attention based multi-scale feature collection and distribution module (SA-MFCD) to fuse the feature maps from different layers of encoders, refine them with spatial attention, and distribute them to different layers of decoders. 
Experimental results on the MS COCO dataset indicate that our networks can remarkably improve the accuracy of human pose estimation over SBN, our network with ResNet34 can even achieve the same accuracy as SBN with ResNet152, and our network can achieve superior results with big backbone networks.


\section{Related~Work}
\subsection{Multi-Person Pose Estimation}
Most multi-person pose estimation approaches can be classified into top-down and bottom-up approaches.

\textbf{Top-Down Multi-Person Pose Estimation.} Top-Down approaches first apply human detection to an image and then predicate the joint locations of each human instance respectively. Convolutional Pose Machine (CPM)~\cite{wei2016convolutional} employs a multi-stage architecture to combine the prediction of previous stages with the input to refine the pose estimation. Newell~\emph{et~al.}~\cite{newell2016stacked} propose a stacked hourglass network to predict heatmap of 2D joint location.
Mask-RCNN~\cite{he2017mask} extends faster-RCNN~\cite{ren2015faster} to support keypoint predication and obtains very competitive results for human pose estimation. CPN~\cite{chen2018cascaded} uses a GlobalNet to localize the “simple” keypoints and a RefineNet to address the “hard” keypoints with online hard keypoint mining. Xiao~\emph{et~al.}~\cite{xiao2018simple} propose the simple baseline network for human pose estimation which has achieved state-of-the-art results. Recently, most state-of-the-art results have been achieved by top-down approaches. 

\textbf{Bottom-Up Multi-Person Pose Estimation.} Bottom-Up approaches predict all joints first and then group them into different human poses. Some works~\cite{pishchulin2016deepcut,insafutdinov2016deepercut,insafutdinov2017arttrack} formulate joint grouping via a Linear Program. Cao~\emph{et~al.}~\cite{cao2017realtime} use part affinity fields to encode the relation between different joints to help joint grouping. Newell~\emph{et~al.}~\cite{newell2017associative} use a detection heatmap and a tagging heatmap for supervising the joint detection and grouping, and then group joints with similar tags into an individual person. Papandreou~\emph{et~al.}~\cite{papandreou2018personlab} develop a convolutional network to predict body joints and their relative displacements for joint grouping. Bottom-up approaches suffer from high complexity of joint grouping step and the multi-scale problem in joint detection. Although some efficient grouping methods have been proposed, bottom-up pose estimation approaches still can not win the competition against top-down approaches.

\subsection {Attention Mechanism and Global Context Modeling} 
Attention mechanism is widely used in computer vision, which focuses on important features and suppresses unnecessary ones. It can be divided into two categories in terms of the dimension considered, channel-wise attention and spatial-wise attention. SENet~\cite{hu2018squeeze} uses a bottleneck transform to generate channel attention weights for image classification. 
Zhu~\emph{et~al.}~\cite{zhu2019stacked} propose a channel attention mechanism in their residual attention block to refine features for image super-resolution. CBAM~\cite{woo2018cbam} and BAM~\cite{park2018bam} apply both channel and spatial attention modules to refine features for various visual tasks. Chu~\emph{et~al.}~\cite{chu2017multi} use attention models for human pose estimation and model the attention from different context and resolution.  

Spatial attention mechanism has also been used to model global context. The non-local network~\cite{wang2018non} uses a query-specific spatial attention map for each query position to aggregate context features. GCNet~\cite{cao2019gcnet} achieves more efficient global context modeling by using a query-independent attention map to generate a global context feature for all query positions and further re-calibrating the global context feature with channel attention.

\section{Methodology}
\begin{figure*}[!t]
\includegraphics[width=\textwidth]{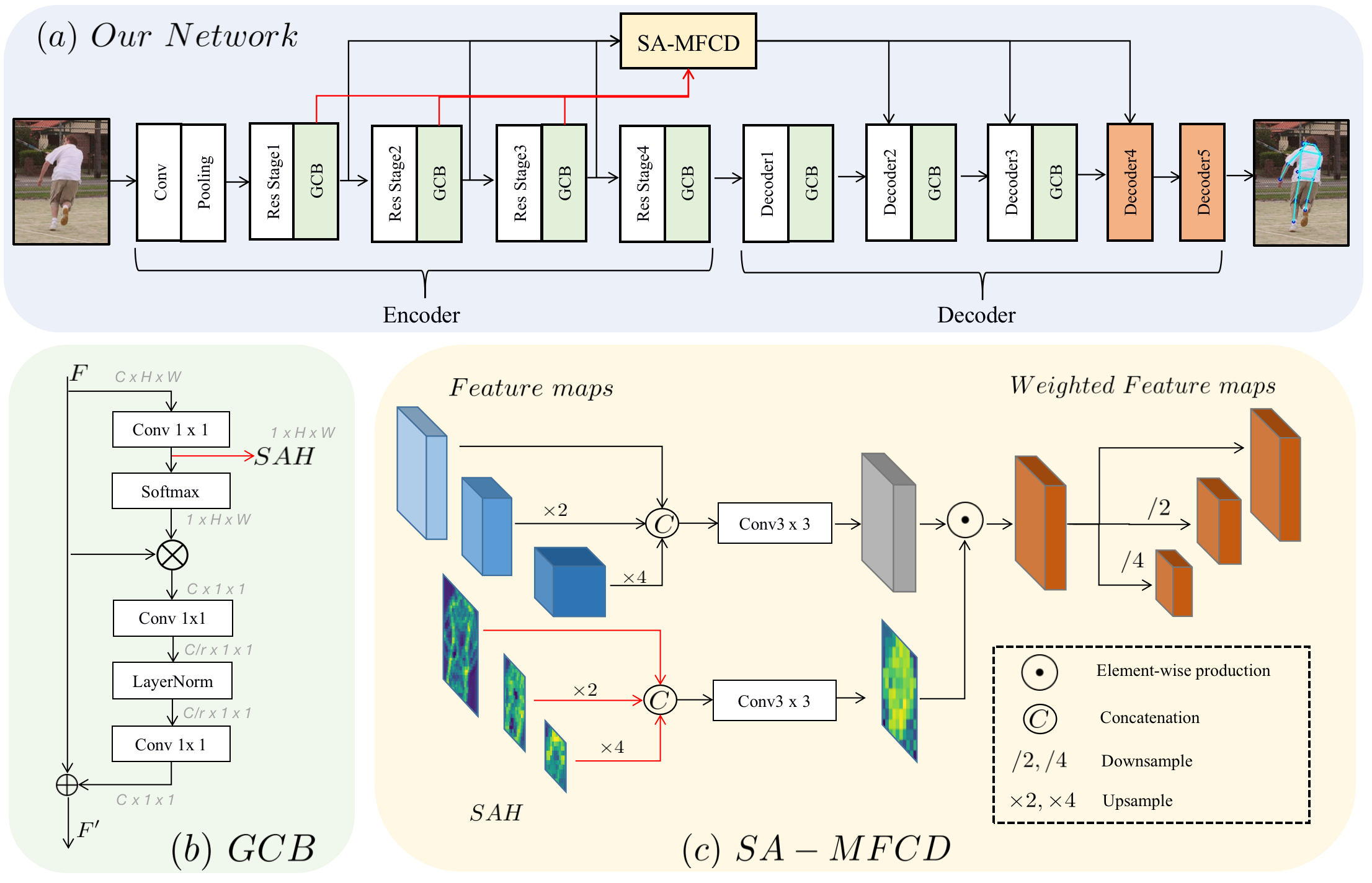}
\caption{(a) The full-resolution encoder-decoder network. The red arrow lines stand for the flows of spatial attention heatmaps (SAH) generated by GCBs, and black arrow lines are for feature maps. (b) The global context block (GCB). $F$ and $F^{\prime}$ are the input and output feature maps of a GCB respectively. (c) The SA-MFCD Module. } \label{fig:2}
\end{figure*}

Our multi-person pose estimation approach uses faster-RCNN~\cite{ren2015faster} to detect each person in an image, and then predicates pose for each person instance. Our network is an extension of the successful simple baseline network (SBN), as show in Fig.~\ref{fig:2}(a). Firstly, two more decoder modules are appended to the end of the SBN to increase the output resolution. The encoder and decoder modules are then enhanced by the global context blocks (GCBs)~\cite{cao2019gcnet} which compute global context features to enhance the original feature maps. Furthermore, a spatial attention based multi-scale feature collection and distribution (SA-MFCD) module is proposed to fuse feature maps from different layers of encoders, refine them with spatial attention and distribute them to different layers of decoders. More detailed information are given in the following subsections.

\subsection{Extend Simple Baseline Network to Full Output Resolution}
The encoder-decoder network structure is widely used in dense prediction, including semantic segmentation~\cite{ronneberger2015u,fu2019stacked}, object detection~\cite{lin2017feature}, and human pose estimation~\cite{newell2016stacked,xiao2018simple}. The encoder-decoder networks contain encoders that gradually reduce the resolutions of feature maps while capturing higher semantic information, and decoders that gradually recover the spatial resolutions. The output resolutions of most popular encoder-decoder networks for human pose estimation are smaller than their input resolutions. For example, the output resolutions of the Hourglass network ~\cite{newell2016stacked} and the simple baseline network ~\cite{xiao2018simple} are only 1/4 of their input resolution. The outputs are then resized to the input resolution by simple transformation, which introduces quantization errors. In~\cite{huang2019devil}, the common biased data processing for human pose estimation has been quantitatively analyzed, and it is found that the error of prediction is mainly due to the horizontal flip operation which is also related to the output stride size. 

To alleviate this problem, we append two more decoder modules to the end of SBN to achieve the full output resolution. Following SBN, the encoder part of our network is a modified version of the ResNet~\cite{he2016deep}, where the average pooling layer and last fully-connected layers are removed and only the convolutional layers are used. It includes a convolutional layer and a pooling layer followed by four stages of ResBlocks, both the first convolutional layer and pooling layer halves the resolution, and at the beginning of each Res-stage (except for the first one), the feature map resolution is also halved by a convolutional layer with strides=2, while the number of filters is doubled. Each decoder is composed of a Deconvolution (Transport Convolution) layer followed by batch normalization and ReLU. The decoder part of our network includes five decoder modules each of which double the resolution, and finally gets output with the same resolution as that of input. Compared to SBN, the two additional decoder modules bring more parameters and increase the model size. Our previous study indicates that reducing the channel sizes of the two additional decoder modules can make a good trade-off between the prediction accuracy and the model size. In our study, we set the channel size of the 4th decoder to 128 and that of the 5th decoder to 32 respectively. We name this basic network as the full resolution version of SBN (FR-SBN).

\subsection{Enhance Encoders and Decoders by Global Context Blocks}

The combination of global context information has proven useful to various visual recognition tasks. However, a convolution layer only models pixel relationship in a local neighborhood, and the long-range dependencies are mainly modeled by deeply stacking convolution layers. Non-local operation~\cite{wang2018non} is proposed to model the long-range dependencies using self-attention mechanism~\cite{vaswani2017attention}. The non-local network uses a query-specific attention map for each query position to aggregate context features and adds the context features to the feature of the corresponding query position. GCNet~\cite{cao2019gcnet} is a more effective and efficient global context modeling approach, which uses a query-independent attention map to generate a global context feature and add it to all query positions.

In this paper, we apply the global context block (GCB)~\cite{cao2019gcnet} to enhance the encoder and decoder modules. GCBs are introduced after the four Res-stages in the encoder part and the first three Decoder Models, as shown in Fig.~\ref{fig:2}(a). The structure of GCB is shown in Fig.~\ref{fig:2}(b). A $1 \times 1$ convolution is first used to generate a spatial attention heatmap (SAH) and followed by softmax to generate attention weights. Then attention pooling is used to obtain a global context feature. After that, a bottleneck transform is applied to it to capture its channel-wise dependencies. Finally, the global context feature is added to features of all positions. The SAH is also used in SA-MFCD which is introduced in the next subsection.


\subsection{Spatial Attention Based Multi-Scale Feature Collection and Distribution}
In previous encoder-decoder networks such as U-Net~\cite{ronneberger2015u} and Hourglass~\cite{newell2016stacked}, skip connections are used to preserve spatial information at each resolution. Skip connections are formed between the encoders and the decoders having the same output resolution to transfer the spatial information across the network for better localization. Through the skip connections, feature maps from the encoder are concatenated (or summed) with the corresponding feature maps of the decoder. For simplicity, the simple baseline network ~\cite{xiao2018simple} does not use skip connections, which may limited its performance.

Different from the traditional skip connection, we propose a Spatial Attention based Multi-Scale Feature Collection and Distribution (SA-MFCD) module which merges multi-scale feature-maps from different layers of encoders, uses spatial attention to refine them, and finally distributes them to different layers of decoders. This scheme can bring multi-scale spatial and semantic information to the decoders for recovering high-resolution feature-maps. The details of this module is show in Fig.~\ref{fig:2}(c). This module has two different inputs: feature maps and spatial attention maps. The feature maps generated from the GCBs after Res-stage1, Res-stage2 and Res-stage3 are sent to SA-MFCD, the feature maps with lower resolutions are up-sampled to match the highest one and concatenated together, and fused by a $3\times 3$ convolution. The SAHs generated from the three GCBs are also resized and concatenated, and fused by a $3\times 3$ convolution. The fused feature maps are then multiplied element-wise with the fused SAH. Finally, the refined feature maps are distributed to Decoder2, Decoder3 and Decoder4 after being resized to corresponding resolutions. The refined feature maps from SA-MFCD are concatenated with the output feature maps of the deconvolution layers in the corresponding decoders and fused by a $3\times 3$ convolution to generate the outputs of the decoders. All of the convolution operations used here are followed by a batch normalization layer and ReLU function.


Via SA-MFCD, feature maps with different resolutions are fused together. The higher semantic information in lower resolution feature maps is combined with the more refined spatial information in the higher resolution feature maps. The spatial attention heatmaps (SAHs) with different resolutions can help extracting useful details and suppressing background information. And the fused feature maps are refined by the spatial attention mechanism to emphasize the positions related to pose estimation. When combining the refined feature maps from SA-MFCD model, the capacities of the corresponding decoders and thus the whole network are improved.


\subsection{Ground-truth and Loss Function}
Our human pose estimation network does not directly predict the coordinates, but do a pixel-level prediction by transforming joint position to heatmap. For each target-person patch cropped by human detection bounding-box, there might exist non-target person in it, we only generate ground-truth for the target one, by a 2D Gaussian function: 
\begin{equation}
\boldsymbol{H}_{k}(x, y)= \exp \left(\frac{-\left[\left(x-x_{k}\right)^{2}+\left(y-y_{k}\right)^{2}\right]}{2 \sigma^{2}}\right),
\end{equation}
where $H_{k}$ is the heatmap for the $k$-th joint$(k \in \left\{1,...,K\right\})$, $(x_k,y_k)$ is the coordinations of the $k$-th joint, $(x,y)$ specifies a pixel location in the heatmap and the hyper-parameter $\sigma$ denotes a pre-fixed spatial variance, which is set to 8 and 12 for $256\times192$ and $388\times288$ input image size respectively. 

As in ~\cite{xiao2018simple,fang2017rmpe}, we also choose Mean-Squared Error (MSE) as the loss function. The MSE loss function is given as:
\begin{equation}
\mathcal{L}_{\mathrm{mse}}=\frac{1}{K} \sum_{k=1}^{K}\left\|\boldsymbol{H}_{k}-\hat{\boldsymbol{H}}_{k}\right\|_{2}^{2},
\end{equation}
where $\hat{H}_{k}$ refers to the predicted heatmap for the $k$-th joint.

\section{Experiments}
\subsection{Experiment Setup}
\textbf{Dataset.}~\emph{MS COCO} keypoint dataset~\cite{lin2014microsoft} has more than 200k images in the wild, where more than 250k person instances are labeled with 17 human joints. We train our networks on the COCO 2017 training dataset, which has 57k images and 150k labeled person instances, no extra data involved. We evaluate our networks on the val2017 set (5k images) and test-dev2017 set (20k images) respectively.

\textbf{Performance Metrics.} 
The OKS-based average precision and recall scores$\footnote{https://cocodataset.org/\#keypoints-eval.}$ are used to evaluate the accuracy of keypoint localization. The object keypoint similarity (OKS) is defined as the distance between the predicted points and the ground-truth normalized by scale of the person. The mean average precision (AP) over 10 OKS thresholds is used as the main evaluation metric. In our experiments, we report AP (the mean of AP scores at OKS = 0.50, 0.55, ..., 0.90, 0.95), AP$^{50}$ (AP at OKS = 0.50), AP$^{75}$, AP$^M$ for medium objects, AP$^L$ for large objects, and AR (the mean of recalls).

\textbf{Implementation Details.} We implemented our networks via PyTorch~\cite{paszke2019pytorch} deep learning framework. ResNet~\cite{he2016deep} pre-trained on ImageNet is used to construct the encoder part in our network. The base learning rate is 1e-3, droped by 0.1 at step 90 and 120, with 140 epochs in total. We utilize Adam optimizer for training. 
Data augmentation is a very important training strategy, we adopt random horizontal flip, random rotation (-40 to +40) degrees and random scale (0.7 to 1.3).

\subsection{Ablation Study}
We first study the effectiveness of each component used to extend SBN, including the two more decoder modules appended to SBN (FR-SBN), the GCBs used to enhance encoders and decoders, and our SA-MFCD module. All versions of networks use ResNet34 to build their encoder part. We evaluate these versions of networks on the COCO val2017 set, all results are obtained over the input size of $256\times 192$ and the same data augmentation strategies.

\begin{table}[!t]
\centering
\caption{Comparison of networks with different additional components}
\begin{tabular}{|c|c|}
\hline
Method	&AP\\
\hline

SBN-34&	69.8\\
\hline
FR-SBN&	71.0\\
\hline
FR-SBN+GCB&	71.6\\
\hline
FR-SBN+GCB+skip & 71.7\\
\hline
FR-SBN+GCB+SA-MFCD	&\textbf{72.5}\\
\hline
\end{tabular}
\label{tablea4}
\end{table}

Table~\ref{tablea4} shows the comparison results in detail. Our re-trained SBN-34 achieved an AP of 69.8. Comparing FR-SBN with SBN-34 (71.0 vs. 69.8), it can be seen that increase of the output resolution effectively improves AP by 1.2 points, which may due to the reduced quantization error. It can also be found that GCB can improve 0.6 AP points (71.6 vs. 71.0) over FR-SBN. SA-MFCD can further improve AP by 0.9 points (72.5 vs. 71.6), while the vanilla skip-connection (FR-SBN+GCB+skip) can only improve AP by 0.1 points (71.7 vs. 71.6). In total, our network (FR-SBN+GCB+SA-MFCD) gets 2.7 AP point improvements over the baseline (SBN-34).


\subsection{Comparison with the state of the art methods}

\begin{table*}[!t]
\centering
\caption{Comparisons on the COCO validation set. OHKM = online
hard keypoints mining}
\begin{tabular}{|c|c|c|c||c|c|c|}

\hline
Method &	Backbone&	Input size &AP &\#Params&GFLOPs&Time \\
\hline
\hline
Hourglass~\cite{newell2016stacked}	&-&$256\times 192$	&66.9&25M&14.3G&-\\
CPN~\cite{chen2018cascaded}	&Resnet50&$256\times 192$		&68.7&27M&6.2G&-\\
CPN+OHKM~\cite{chen2018cascaded}	&Resnet50	&$256\times 192$	&69.4&27M&6.2G&-\\
PoseFix~\cite{moon2019posefix} & CPN &$256\times 192$	&72.1 &-&-&-\\
SBN-50~\cite{xiao2018simple}&	Resnet50&$256\times 192$	&	70.4&34M&8.9G&2.3ms\\
SBN-101~\cite{xiao2018simple}	&Resnet101&$256\times 192$	&	71.4&53M&12.4G&3.3ms\\
SBN-152~\cite{xiao2018simple}	&Resnet152	&$256\times 192$	&72.0&68.6M&15.7G&4.8ms\\
\hline
\textbf{ours}&	Resnet34&$256\times 192$	&	\textbf{72.5}&33M&21.3G&3.5ms\\
\hline
\hline
SBN-50~\cite{xiao2018simple}&	Resnet50&$384\times 288$	&	72.2&34M&20G&5.7ms\\
SBN-101~\cite{xiao2018simple}	&Resnet101&$384\times 288$	&	73.6&53M&27.9G&8.0ms\\
SBN-152~\cite{xiao2018simple}	&Resnet152	&$384\times 288$	&\textbf{74.3}&68.6M&35.3G&11.8ms\\
\hline
\textbf{ours}&	Resnet34&$384\times 288$	&	74.2&33M&48.0G&8.6ms\\
\hline

\end{tabular}
\label{tablea5}
\end{table*}
\begin{table*}[!t]
\centering
\caption{Comparisons on the COCO test-dev set. $^{*}$ means use extra data}
\begin{tabular}{|c|c|c|c|cccc|c|}

\hline
Method &	Backbone&	Input size &AP &AP$^{50}$ &AP$^{75}$ &AP$^{M}$ &AP$^{L}$ &AR \\
\hline
\hline
Mask.~\cite{he2017mask}&Res.50-FPN&-&63.1& 87.3& 68.7& 57.8& 71.4& -\\
Integral.~\cite{sun2018integral}&Res.101&$256 \times 256$&67.8 &88.2& 74.8& 63.9& 74.0 &-\\
CSANet~\cite{yu2019context} &Res.101&$256\times 192$&72.3 &91.2 &80.2 &69.3 &77.6 &79.1\\
CASNet~\cite{yu2019context} &Res.152&$256\times 192$&72.8&91.4&80.9&69.8&78.3&79.6\\
UDP\cite{huang2019devil}&Res.152&$256\times 192$&72.9&91.6&80.9&70.0&78.5&78.4\\
HRNet\cite{sun2019deep}&HRNet-w48&$256\times 192$&\textbf{74.3}&\textbf{92.4}&\textbf{82.6}&\textbf{71.2}&\textbf{79.6}&\textbf{79.7}\\
\hline
\textbf{ours}&Res.34&$256 \times 192$&72.0&91.1&79.9&68.9&77.7&77.5\\
\textbf{ours}&Res.50&$256 \times 192$&72.3&91.3&80.1&69.1&78.2&77.8\\
\textbf{ours}&Res.101&$256 \times 192$&72.6&91.6&80.5&69.4&78.4&78.1\\
\textbf{ours}&Res.152&$256 \times 192$&73.3&91.3&80.8&70.1&78.9&78.8\\
\hline
\hline
CPN~\cite{chen2018cascaded}&Res.Inc.&$384 \times 288$&72.1 &91.4& 80.0& 68.7& 77.2& 78.5\\

CFN~\cite{huang2017coarse}&-&-&72.6 &86.1& 69.7& 78.3& 64.1& -\\
CPN (ens.)~\cite{chen2018cascaded}&Res.Inc. &$384 \times 288$& 73.0 &91.7& 80.9& 69.5& 78.1& 79.0\\
SBN~\cite{xiao2018simple}&Res.152 &$384 \times 288$&73.7 &91.9& 81.1& 70.3& 80.0 &79.0\\
PoseFix~\cite{moon2019posefix}&Res.152&$384 \times 288$&73.6&90.8&81.0&70.3 &79.8& 79.0\\
CSANet~\cite{yu2019context} &Res.50&$384 \times 288$&73.5 &91.4 &80.8 &69.9 &79.4 &79.7\\

CSANet~\cite{yu2019context} &Res.101&$384 \times 288$&74.1 &91.6 &81.6 &70.7 &79.8 &80.4\\
CSANet~\cite{yu2019context} &Res.152&$384 \times 288$&74.5 & 91.7 &82.1 &71.2 &80.2 &80.7\\
HRNet~\cite{sun2019deep} &HRNet-w48&$384 \times 288$&\textbf{75.5} & \textbf{92.5} &\textbf{83.3} &\textbf{71.9} &\textbf{81.5} &\textbf{80.5}\\
UDP~\cite{huang2019devil} &Res.50&$384 \times 288$ &72.5&91.1 &79.7&68.8& 79.1 &77.9\\
UDP~\cite{huang2019devil} &Res.152&$384 \times 288$ &74.7&91.8 &82.1&71.5& 80.8 &80.0\\
\hline

\textbf{ours}&Res.34&$384 \times 288$&73.7&91.7&81.1&70.4&79.7&78.9\\

\textbf{ours}&Res.50&$384 \times 288$&73.9&91.7&81.3&70.4&80.0&79.0\\

\textbf{ours}&Res.101&$384 \times 288$&74.3&92.0&82.0&71.0&80.4&79.6\\

\textbf{ours}&Res.152&$384 \times 288$&74.9&92.2&82.4&71.6&80.9&80.2\\
\hline
\end{tabular}
\label{tablea51}
\end{table*}

Our method is compared to the state-of-the-art methods on the validation and test set with two different input sizes respectively. Most of the compared methods are built on ResNet. PoseFix~\cite{moon2019posefix} uses CPN as the backbone, and UDP~\cite{huang2019devil} uses SBN as the backbone. Table~\ref{tablea5} gives the results on the validation set. It can be found that our method achieves the best results when the input size is $256\times 192$. When the input size rises to $384\times 288$, our network with ResNet34 can still achieve almost the same AP of SBN-152 (74.2 vs. 74.3) with only half of parameters and less inference time (8.6ms vs. 11.8ms), but the GFLOPs of our network is higher than that of SBN-152. This is because that our backbone is relatively small and the SA-MFCD module can run in parallel with the encoder-decoder network. The experimental results on the validation dataset indicate our network with ResNet34 can achieve the same accuracy as SBN with ResNet152, but with less inference time.

Table~\ref{tablea51} gives the results on test-dev2017 set. When using the input of $256\times 192$, our network outperforms CSANet~\cite{yu2019context} and UDP\cite{huang2019devil} with the same backbone networks. When using the input of $384\times 288$, our network with ResNet34 outperforms CSANet and UDP with ResNet50. When ResNet50 is used as the backbone, our network is 0.4 points higher than CSANet and 1.4 points higher than UDP. When using ResNet101, our network reaches higher results than CSANet (74.3 vs. 74.1). When ResNet152 is used, our results (74.9) are also better than UDP (74.7) and CSANet (74.5). HRNet is the method outperforming our network at both input size settings, which is a more advanced and elaborate network recently refreshing the state-of-the-art of many computer visual tasks.

\section{Conclusions and Future Works}
This paper has proposed an advanced encoder-decoder networks for human pose estimation, which improves the effectiveness of the simple baseline network. To the best of our knowledge, we are the first to develop network with full output resolution for human pose estimation, which can reduce the quantization error and recover more accurate spatial information. Global context blocks are also used to enhance the encoders and decoders. Furthermore, the proposed spatial attention based multi-scale feature collection and distribution module can fuse multi-scale feature information to boost the pose estimation. Our experimental results show that our networks can remarkably improve the prediction accuracy over SBN, and our network with big backbone network can achieve top ranked performance among recent works. In this paper, our network uses ResNet to build the encoder part, and it is worth trying some more advanced backbones in the future. 
\section{Acknowledgement}
This work was supported in part by the National Natural Science Foundation of China (U20B2063), the Sichuan Science and Technology Program, China (2020YFS0057), and the Fundamental Research Funds for the Central Universities (ZYGX2019Z015).

\bibliographystyle{ACM-Reference-Format}
\bibliography{acmart}

\end{document}